\documentclass[10pt,twocolumn,letterpaper]{article}

\usepackage{cvpr}
\usepackage{times}
\usepackage{epsfig}
\usepackage{graphicx}
\usepackage{amsmath}
\usepackage{amssymb}
\usepackage{microtype}

\usepackage[pagebackref=true,breaklinks=true,colorlinks,bookmarks=false]{hyperref}
\usepackage{amsfonts,amsthm,mdframed,subcaption,algpseudocode,algorithm,color,tikz,mdframed}
\usetikzlibrary{positioning,decorations.pathreplacing,calc,arrows}

\def\cvprPaperID{3235}
\ifcvprfinal\fi

\def\nsim{{\:\sim\:}}
\let\citep=\cite
\let\citet=\cite

\newtheorem{hypothesis}{Hypothesis}
\newcommand{\R}{\mathbb{R}}

\cvprfinalcopy

\tikzset{
  >=stealth',
  cir/.style={
    rectangle,
    rounded corners,
    draw=black, thick,
    minimum width=1.5em,
    text centered},
  arr/.style={
    ->,
    thick,
    shorten <=2pt,
    shorten >=2pt,},
  layer/.style={
    draw=black,
    fill=white,
    thick,
    rectangle,
    rounded corners,
    minimum width=1.5em,
    minimum height=12em
  },
  brace/.style={
    thick,
    decoration={
    brace,
    mirror,
    raise=2.5cm
    },
    decorate
  }
}

\newcommand{\drawlayer}[4]{
  \node[layer, draw=#4] (#1) at (#2,#3) {};
  \foreach \x in {-1.75, -1.25, -0.75, -0.25, 0.25, 0.75, 1.25, 1.75} 
    \draw[thick, #4] (#2,\x+#3) circle (0.15);
}

\begin{document}

\title{Discovering Causal Signals in Images}

\author{
  David Lopez-Paz\\
  Facebook AI Research\\
  {\tt\small dlp@fb.com} \\
  \and
  Robert Nishihara\\
  UC Berkeley\\
  {\tt\small rkn@eecs.berkeley.edu} \\
  \and
  Soumith Chintala\\
  Facebook AI Research\\ 
  {\tt\small soumith@fb.com} \\
  \and
  Bernhard Sch\"olkopf\\
  MPI for Intelligent Systems\\
  {\tt\small bs@tue.mpg.de} \\
  \and
  L\'eon Bottou\\
  Facebook AI Research\\
  {\tt\small leon@bottou.org} \\
}

\maketitle

\begin{abstract}%
  This paper establishes the existence of observable footprints 
  that reveal the ``{causal dispositions}'' of the object
  categories appearing in collections of images.  We achieve this
  goal in two steps.  First, we take a learning approach to
  observational causal discovery, and build a classifier that achieves
  state-of-the-art performance on finding the causal direction between
  pairs of random variables, given samples from their joint
  distribution.  Second, we use our causal direction classifier to
  effectively distinguish between features of objects and features of
  their contexts in collections of static images.  Our experiments
  demonstrate the existence of a relation between the direction of
  causality and the difference between objects and their contexts, and
  by the same token, the existence of observable signals that reveal
  the causal dispositions of objects.
\end{abstract}

\section{Introduction}\label{sec:intro}
\begin{figure}
  \includegraphics[width=\linewidth]{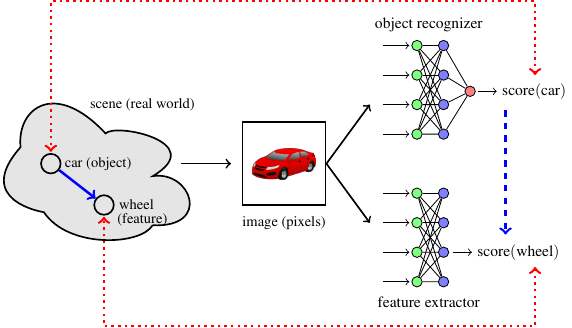}
  \caption{Our goal is to reveal causal relationships between pairs of
    \emph{real entities} composing scenes in the world (\eg ``the presence of
    cars cause the presence of wheels'', solid blue arrow).
    To this end, we apply a novel observational
    causal discovery technique, NCC, to the joint distribution of a
    pair of related \emph{proxy variables} that are computed by
    applying CNNs to the image pixels. Since these variables are
    expected to be highly correlated with the presence of the
    corresponding real entities, the appearance of causation between
    the proxy variables (dashed blue arrow) suggests that there is a causal
    link between the real world entities themselves (\eg. the appearance of
    causation between \emph{score(car)} and \emph{score(wheel)} suggests that the
    presence of cars causes the presence of wheels in the real world.)
  }
  \label{fig:aliasing}
\end{figure}

Imagine an image representing a bridge over a river.  On top of the bridge, a
car is speeding through the right lane.
 
Modern computer vision algorithms excel at answering questions about the
observable properties of the scene, such as such as {\emph{``Is there a car in
this image?''}}.  This is achieved by leveraging correlations between pixels
and image features across large datasets of images. However, a more nuanced
understanding of images arguably requires the ability to \emph{reason about}
how the scene depicted in the image would change in response to interventions.
Since the list of possible interventions is long and complex, we can, as a
first step, reason about the intervention of removing an object.

To this end, consider the two counterfactual questions \emph{``What would the
scene look like if we were to remove the car?''} and \emph{``What would the
scene look like if we were to remove the bridge?''} On the one hand, the first
intervention seems rather benign. After removing the car, we could argue that
the rest of the scene depicted in the image (the river, the bridge) would
remain invariant. On the other hand, the second intervention seems more severe.
If the bridge had been removed from the scene, it would, in general, make
little sense to observe the car floating weightlessly above the river. Thus, we
understand that the presence of the bridge has an \emph{effect} on the presence
of the car.  Reasoning about these and similar counterfactuals allows to begin
asking \emph{``Why is there a car in this image?''} This question is of course
poorly defined, but the answer is linked to the causal relationship between the
bridge and the car. In our example, the presence of the bridge {\em causes} the
presence of the car, in the sense that if the bridge were not there, then the
car would not be either (needless to say, it is not the {\em only} cause for
the car). Such \emph{interventional semantics} of what is meant by {\em
causation} align with current approaches in the literature
\cite{rubin-1986,Pearl00}.

\subsection{Causal dispositions}
\label{sec:causaldispositions}
We have so far discussed causal relations between two objects present in a
single image, representing a particular scene. In order to deploy statistical
techniques, we must work with a large collection of images representing a
variety of scenes.  Similar objects may have different causal relationships in
different scenes. For instance, an image may show a car passing under the
bridge, instead of over the bridge.

The \emph{dispositional semantics} of causation \cite{mumford-anjum-2011}
provide a way to address this difficulty. In this framework, causal relations
are established when objects exercise some of their \emph{causal dispositions},
which are sometimes informally called the \emph{powers of objects}. For
instance a bridge has the power to provide support for a car, and a car has the
power to cross a bridge.  Although the objects present in a scene do not
necessarily exercise all of their powers, the foundation of the dispositional
theory of causation is that all causal relationships are manifestations of the
powers of objects.\footnote{\relax Causal dispositions are more primitive
concepts than the causal graphs of Pearl's approach \cite{Pearl00}.  Therefore,
in our case, causal dispositions are responsible for the shape of causal
graphs.}

Since the list of potential causal dispositions is as long and complex as the
list of possible interventions, we again restrict our attention to
interventions that affect the presence of certain objects in the scene. In
particular we can count the number $\mathcal{C}(A,B)$ of images in which the
causal dispositions of objects of categories A and B are exercised in a manner
that the objects of category B would disappear if one were to remove objects of
category A. We then say that the objects of category A \emph{cause the
presence} of objects of category B when $\mathcal{C}(A,B)$ is (sufficiently)
greater than the converse $\mathcal{C}(B,A)$.  This definition induces a
network of asymmetric causal relationships between object categories that
represents, on average, how real-world scenes would be modified when one were
to make certain objects disappear.

The fundamental question addressed in this paper is to
determine whether such an asymmetric causal relationship
can be inferred from statistics observed in image datasets.

\begin{hypothesis}\label{hyp:thereissignal}
  Image datasets carry an observable statistical signal
  revealing the asymmetric relationship between object categories
  that results from their causal dispositions.
\end{hypothesis}

To our knowledge, no prior work has established or even considered the
existence of such a signal.  If such a signal were found, it would imply that
it is in principle possible for \emph{statistical computer vision algorithms to
reason about the causal structure of the world.} This is not small feat, given
that it is being debated in statistics until this day whether one can at all
infer causality from purely statistical information, without performing
interventions.  The focus of this contribution is to establish the existence of
such causal signals using a newly proposed method.  We do not, in contrast,
make any engineering contribution advancing the state-of-the-art in standard
computer vision tasks using these signals --- this is beyond the scope of the
present paper.

\subsection{Object features and context features}
\label{sec:hypotheses}
Since image datasets do not provide labels describing the causal dispositions
of objects, we cannot resort to supervised learning techniques to find the
causal signal put forward by Hypothesis~\ref{hyp:thereissignal}.  Instead, we
take an indirect approach described below.

The features computed by the final layers of a 
convolutional neural network (CNN) \cite{lecun-89e,OquabBLS15,KaimingHeZRS16}
often indicate the presence of a well localized object-like feature in the
scene depicted by the image under study.\footnote{\relax
The word \emph{feature} in this work describes a property of the
scene whose presence is flagged by \emph{feature activations}
computed by the CNN.}  Various techniques have been developed to
investigate where these object-like features appear
\textit{in the scene} and what they look like
\textit{in the image} \cite{Zhou,ZeilerF13}.  We can therefore
examine large collections of images representing different
\emph{objects of interest} such as cats, dogs, trains, buses, cars,
and people.  The locations of these objects in the images are given to
us in the form of bounding boxes. For each object of interest, we can
distinguish between \emph{object features} and \emph{context
  features}.  By definition, the object features are those that are mostly
activated inside the bounding box of the object of interest, and
the context features are those that are mostly activated outside the bounding
box of the object of interest.
Independently and in parallel, we also distinguish
between \emph{causal features} and \emph{anticausal features}
\cite{Scholkopf12}.  Causal features are those that \emph{cause the
presence{\footnotemark} of the object in the scene}, whereas
anticausal features are those \emph{caused by the presence of the
object in the scene}.  \footnotetext{In the sense defined in
Section~\ref{sec:causaldispositions}.}

Having made a distinction between object and context features, our indirect
approach relies on a second hypothesis:
\begin{hypothesis}\label{hyp:our-hypothesis}
  There exists an observable statistical dependence between object features and
  anticausal features.  The statistical dependence between context features and
  causal features is nonexistent or much weaker.
\end{hypothesis}

We expect Hypothesis~\ref{hyp:our-hypothesis} to be true, because many
of the features caused by the presence of an object of interest are
in fact parts of the object itself, and hence are likely to be contained inside
its bounding box. For instance, the presence of a car often causes the
presence of wheels.  In contrast, the context of an object of interest
may either cause or be caused by the presence of the object. For
instance, asphalt-like features cause the presence of a car, but the 
car's shadow is caused by the presence of the car. Importantly, empirical
support in favour of Hypothesis~\ref{hyp:our-hypothesis} translates
into support in favour of Hypothesis~\ref{hyp:thereissignal}.

\subsection{Our contribution}
\label{sec:our-contribution}

Our plan is to use a large collection of images to provide empirical evidence
in favour of Hypothesis~\ref{hyp:our-hypothesis}.
In order to do so, we must effectively determine,
for each object category, 
which features are causal or anti-causal. In this manner we would support 
Hypothesis~\ref{hyp:our-hypothesis}, and consequently, 
Hypothesis~\ref{hyp:thereissignal}.

Our exposition is organized as follows.  After a discussing related literature,
Section~\ref{sec:cause-effect} introduces the basics of causal inference from
observational data.  Section~\ref{sec:ncc} proposes a new algorithm, the Neural
Causation Coefficient (NCC), able to learn causation from a corpus of labeled
data. NCC is shown to outperform the previous state-of-the-art in cause-effect
inference.  Section~\ref{sec:visual-causation} makes use of NCC to distinguish
between causal and anticausal features in collections of images. As
hypothesized, we show a consistent relationship between anticausal features and
object features.  Finally, Section~\ref{sec:conclusion} closes our exposition
by offering some conclusions and directions for future research. 

\subsection{Related work}

The experiments described in this paper depend crucially on the properties of
the features computed by the convolutional layers of a CNN \cite{lecun-89e}.
Zeiler \etal \cite{ZeilerF13} show that the final convolutional layers can
often be interpreted as object-like features. Work on weak supervision
\cite{OquabBLS15,Zhou} suggests that such features can be accurately localized.

We also build on the growing literature discussing the discovery of causal
relationships from observational data
\cite{HoyJanMooPetetal09,Mooij14,Lopez-Paz15a,chalupka2016estimating}.  In
particular, the Neural Causation Coefficient (Section~\ref{sec:ncc}) is related
to \cite{Lopez-Paz15a} but offers superior performance, and is learned
end-to-end from data. The notion of causal and anticausal features was inspired
by \cite{Scholkopf12}. We believe that our work is the first observational
causal discovery technique that targets the causal dispositions of objects.

Causation in computer vision has been the object of at least four recent works.
Pickup \etal \cite{PickupPWSZZSF14} use observational causal discovery
techniques to determine the direction of time in video playback.  Lebeda \etal
\cite{lebeda2015exploring} use transfer entropy to study the causal
relationship between object and camera motions in video data.  Fire and Zhu
\cite{fire2013,fire2016learning} use video data annotated with object status
and actions to infer perceptual causality.  The work of Chalupka \etal.
\cite{ChalupkaPE15} is closer to our work because it addresses causation issues
in images. However, their work deploys \emph{interventional} experiments to
target causal relationships in the labelling process, that is, which pixel
manipulations can result in different labels, whereas we target causal
relationships in scenes from a purely observational perspective. This critical
difference leads to very different conceptual and technological challenges. 

\section{Observational causal discovery}
\label{sec:cause-effect}

Randomized experiments are the gold standard for causal inference
\citep{Pearl00}. Like a child may drop a toy to probe the nature of
gravity, these experiments rely on interacting with the world to reveal 
causal relations between variables of interest.  When such experiments
are expensive, unethical, or impossible to conduct, we must discern
cause from effect using observational data only, and without the
ability to intervene \citep{steyvers2003inferring}. This is the domain
of \emph{observational causal discovery}.

In the absence of any assumptions, the determination of causal
relations between random variables given samples from their joint
distribution is fundamentally impossible 
\citep{Pearl00,PetMooJanSch14}.  However, it may still be possible to
determine a plausible causal structure in practice.  For joint
distributions that occur in the real world, the different causal
interpretations may not be equally likely. That is, the causal
direction between typical variables of interest may leave a detectable
signature in their joint distribution. We shall exploit this insight
to build a classifier for determining the cause-effect relation
between two random variables from samples of their joint distribution.

In its simplest form, observational causal discovery
\citep{PetMooJanSch14,Mooij14,Lopez-Paz15b} considers the \emph{observational
sample}
\begin{align}
  \label{eq:observation}
  S &= \{ (x_j, y_j) \}_{j=1}^m \sim P^m(X,Y),
\end{align}
and aims to infer whether $X \to Y$ or $Y \to X$. In particular, $S$ is assumed
to be drawn from one of two models: from a \emph{causal model} where $X \to Y$,
or from an \emph{anticausal model} where $X \leftarrow Y$.
Figure~\ref{fig:models} exemplifies a family of such models, the Additive Noise
Model (ANM) \citep{HoyJanMooPetetal09}, where the effect variable $Y$ is a
nonlinear function $f$ of the cause variable $X$, plus some independent random
noise $E$.

\begin{figure}
  \begin{center}
  \begin{minipage}{4.1cm}
    \begin{algorithmic}
    \State $f \sim P_f$
    \For{$j = 1, \ldots, m$}
      \State $x_j \sim P_c(X)$
      \State $e_j \sim P_e(E)$
      \State $y_j \leftarrow f(x_j) + e_j$
    \EndFor
    \State \Return $S = \{(x_j, y_j)\}_{j=1}^m$
    \end{algorithmic}
  \end{minipage}
  \end{center}
  \caption{Additive Noise Model, where $X \to Y$.}
  \label{fig:models}
\end{figure}

If we make no assumptions about the distributions $P_f$, $P_c$, and $P_e$
appearing in Figure~\ref{fig:models}, the problem of observational causal
discovery is \emph{nonidentifiable} \citep{PetMooJanSch14}.  To address this
issue, we assume that whenever $X \to Y$, the cause, noise, and mechanism
distributions are ``independent''.  This should be interpreted as an informal
statement that includes two types of independences. One is the independence
between the cause and the mechanism (ICM) \citep{Lemeire06,Scholkopf12}, which
is formalized not as an independence between the input variable $x$ and the
mechanism $f$, but as an independence between the data source (that is, the
distribution $P(X)$) and the mechanism $P(Y|X)$ mapping cause to effect. This
can be formalized either probabilistically \citep{DanJanMooZscSteZhaSch10} or
in terms of algorithmic complexity \citep{JanSch10}.  The ICM is one
incarnation of uniformitarianism: processes $f$ in nature are fixed and
agnostic to the distributions $P_c$ of their causal inputs.  The second
independence is between the cause and the noise. This is a standard assumption
in structural equation modeling, and it can be related to causal sufficiency.
Essentially, if this assumption is violated, our causal model is too small and
we should include additional variables \cite{Pearl00}.  In lay terms, believing
these assumptions amounts to not believing in spurious correlations.

For most choices of $(P_c, P_e, P_f)$, the ICM will be violated in the
anticausal direction $X \leftarrow Y$. This violation will often leave an
observable statistical footprint, rendering cause and effect distinguishable
from observational data alone \citep{PetMooJanSch14}. But, what exactly are
these \emph{causal footprints}, and how can we develop statistical tests to
find them?

\subsection{Examples of observable causal footprints}

\begin{figure}
  \begin{center}
  \begin{subfigure}{.2984\linewidth}
    \centering
    \includegraphics[width=\textwidth]{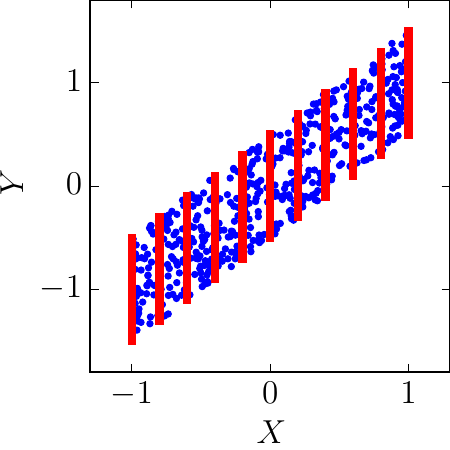}
    \caption{ANM $X \to Y$.}
    \label{fig:footprints:anm:a}
  \end{subfigure}
  \hfill
  \begin{subfigure}{.2984\linewidth}
    \centering
    \includegraphics[width=\textwidth]{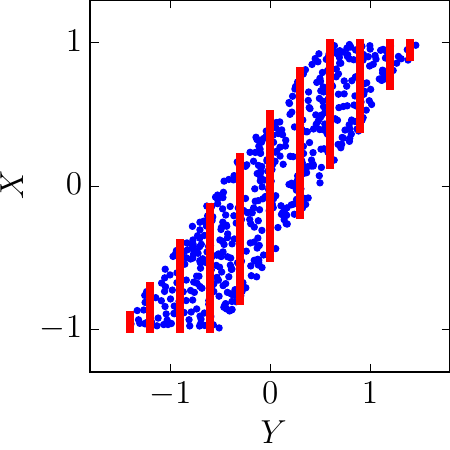}
    \caption{ANM $Y \to X$}
    \label{fig:footprints:anm:b}
  \end{subfigure}
  \hfill
  \begin{subfigure}{.3732\linewidth}
    \centering
    \includegraphics[width=\textwidth]{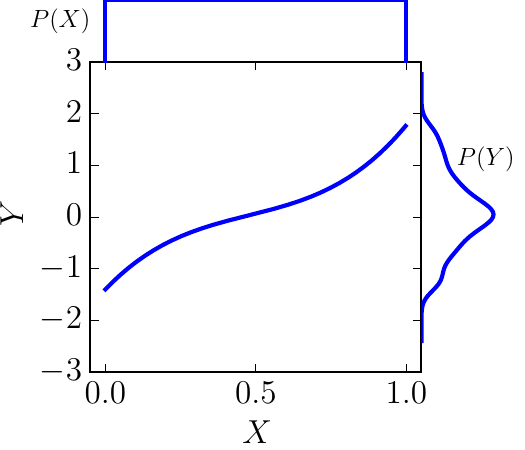}
    \vskip-0.19cm
    \caption{Monotonic $X \to Y$.}
    \label{fig:footprints:igci}
  \end{subfigure}
  \end{center}
  \caption{Examples of causal footprints.}
  \label{fig:footprints}
\end{figure}

Let us illustrate two types of observable causal footprints.

First, consider a linear additive noise model $Y \leftarrow f(X)+E$, where the
cause $X$ and the noise $E$ are two \emph{independent} uniform random variables
with bounded range, and the mechanism $f$ is a linear function
(Figure~\ref{fig:footprints:anm:a}).  Crucially, it is impossible to construct
a linear additive noise model $X \leftarrow \tilde{f}(Y) + \tilde{E}$ where the
new cause $Y$ and the new noise $\tilde{E}$ are two \emph{independent} random
variables (except in degenerate cases). This is illustrated in
Figure~\ref{fig:footprints:anm:b}, where the variance of the new noise variable
$\tilde{E}$ varies (as depicted in red bars) across different locations of the
new cause variable $Y$.  Therefore, the ICM assumption is satisfied for the
correct causal direction $X \to Y$ but violated for the wrong causal direction
$Y \to X$. This asymmetry makes cause distinguishable from effect
\citep{HoyJanMooPetetal09}.  Here, the relevant footprint is the independence
between $X$ and $E$.

Second, consider a new observational sample where $X \to Y$, $Y=f(X)$, and $f$
is a monotone function. The causal relationship $X \to Y$ is deterministic, so
the noise-based footprints from the previous paragraphs are rendered useless.
Let us assume that $P(X)$ is a uniform distribution. Then, the probability
density function of the effect $Y$ increases whenever the derivative $f'$
decreases, as depicted by Figure~\ref{fig:footprints:igci}. Loosely speaking,
the shape of the effect distribution $P(Y)$ is thus not independent of the
mechanism $f$. In this example, ICM is satisfied under the correct causal
direction $X \to Y$, but violated under the wrong causal direction $Y \to X$
\citep{DanJanMooZscSteZhaSch10}. Again, this asymmetry renders the cause
distinguishable from the effect \cite{DanJanMooZscSteZhaSch10}. Here, the
relevant footprint is a form of independence between the density of $X$ and
$f'$.

It may be possible to continue in this manner, considering more classes of
models and adding new footprints to detect causation in each case.  However,
engineering and maintaining a catalog of causal footprints is a tedious task,
and any such catalog will most likely be incomplete. The next section thus
proposes to use neural networks to learn causal footprints directly from data.

\section{The neural causation coefficient}
\label{sec:ncc}
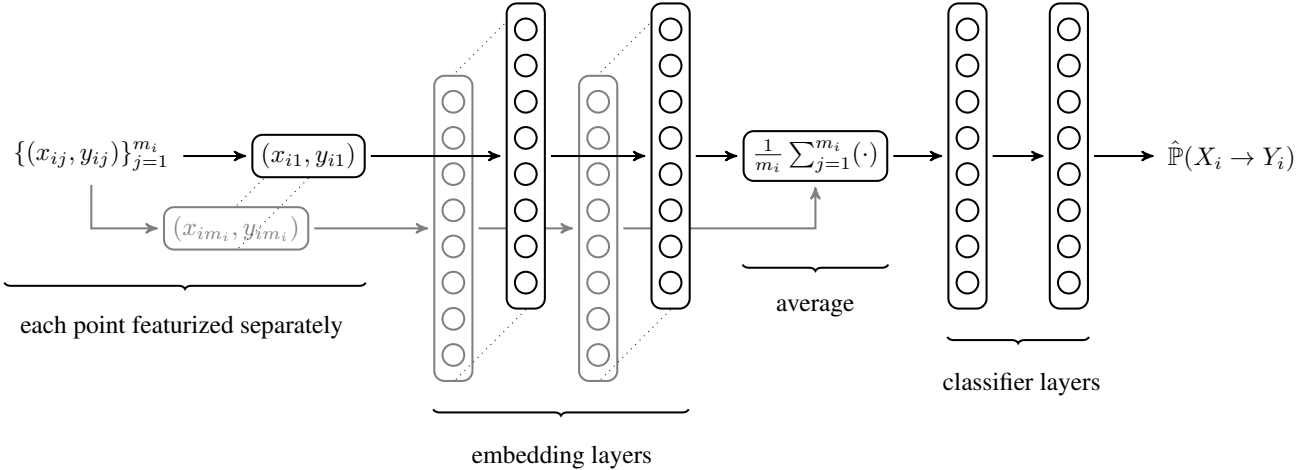
\begin{figure*}[t]
  \begin{center}
  \resizebox{\linewidth}{!}{
  \begin{tikzpicture}
    \node[] (input) at (-6,0) {$\lbrace (x_{ij}, y_{ij}) \rbrace_{j=1}^{m_i}$};
    \node[] (dummy) at (-6,-1) {};
    
    \node[cir] (inp1) at (-3,0) {$(x_{i1}, y_{i1})$};
    \node[cir,gray] (inp2) at (-4,-1) {$(x_{im_i}, y_{im_i})$};
    
    \draw[dotted, shorten <=25pt] (inp1.north) edge (inp2.north);
    \draw[dotted] (inp1.south) edge (inp2.south);
    
    \draw [arr] (input) edge (inp1);
    \path [draw, gray, arr] (input) |- (inp2);

    \drawlayer{emb22}{1}{-1}{gray}
    \node[cir] (mean) at (4,0) {$\frac{1}{m_i} \sum_{j=1}^{m_i} (\cdot)$};
    
    \path [draw, gray, arr] (emb22.east) -| (mean.south);
    \drawlayer{emb21}{2}{0}{}
    
    \drawlayer{emb12}{-1}{-1}{gray}
    \draw [arr,gray] (emb12) edge (emb22);
  
    \drawlayer{emb11}{0}{0}{}
    \draw [arr] (emb11) edge (emb21);
    
    \draw [arr,gray] (inp2) edge (emb12);
    \draw [arr] (inp1) edge (emb11);
    
    \draw[dotted, shorten <=12pt] (emb11.north) edge (emb12.north);
    \draw[dotted] (emb11.south) edge (emb12.south);
    \draw[dotted, shorten <=12pt] (emb21.north) edge (emb22.north);
    \draw[dotted] (emb21.south) edge (emb22.south);
    
    \draw[arr] (emb21) edge (mean);
    
    \drawlayer{clf1}{6.1}{0}{}
    \drawlayer{clf2}{7.5}{0}{}
  
    \node[] (prob) at (9.75,0) {$\hat{\mathbb{P}}(X_i \to Y_i)$};
  
    \draw [arr] (mean) edge (clf1);
    \draw [arr] (clf1) edge (clf2);
    \draw [arr] (clf2) edge (prob);
  
    \draw [brace] (clf1.west) -- (clf2.east);
    
    \draw [brace] (emb12.west) -- (2.25,-1);
  
    \draw [brace, decoration={raise=1.5cm}] (mean.west) -- (mean.east);
    
    \draw [brace, decoration={raise=1.75cm}] (input.west) -- (inp1.east);
    
    \node[below=1.5cm of mean] (meanlabel) {average};
    
    \node[below=0.75cm of clf2, xshift=-.65cm] (clflabel) {classifier layers};
    
    \node[below=0.75cm of emb22, xshift=-.5cm] (clflabel) {embedding layers};
    
    \node[below=1.75cm of input, xshift=1.25cm] (clflabel) {each point featurized separately};
  \end{tikzpicture}
  }
  \end{center}
  \caption{Scheme of the Neural Causation Coefficient (NCC) architecture.}
  \label{fig:ncc}
\end{figure*}

To learn causal footprints from data, we follow \cite{Lopez-Paz15b} and pose
cause-effect inference as a binary classification task. Our input patterns
$S_i$ are effectively scatterplots similar to those shown in
Figures~\ref{fig:footprints:anm:a} and \ref{fig:footprints:anm:b}. That is,
\emph{each data point is a bag of samples $(x_{ij}, y_{ij}) \in \R^2$ drawn iid
from a distribution $P(X_i,Y_i)$}.  The class label $l_i$ indicates the causal
direction between $X_i$ and $Y_i$.
\begin{align}
  D   &= \{(S_i, l_i)\}_{i=1}^n,\nonumber\\
  S_i &= \{(x_{ij},y_{ij})\}_{j=1}^{m_i} \sim P^{m_i}(X_i,Y_i),\nonumber\\
  l_i &= \begin{cases} 0 & \text{if $X_i \to Y_i$} \\ 1 & \text{if $X_i \leftarrow Y_i$} \end{cases}.
  \label{eq:causal-data}
\end{align}
Using data of this form, we will train a neural network to classify samples
from probability distributions as causal or anticausal. Since the input
patterns $S_i$ are not fixed-dimensional vectors, but bags of points, we borrow
inspiration from the literature on kernel mean embedding classifiers
\citep{smola2007hilbert} and construct a feedforward neural network of the form
\begin{equation*}
  \text{NCC}(\{(x_{ij},y_{ij})\}_{j=1}^{m_i}) = \psi\left(\frac{1}{m_i} \sum_{j=1}^{m_i} \phi(x_{ij}, y_{ij})\right).
\end{equation*}
In the previous, $\phi$ is a \emph{feature map}, and the average over all
$\phi(x_{ij}, y_{ij})$ is the \emph{mean embedding} of the empirical
distribution $\frac{1}{m_i} \sum_{i=1}^{m_i} \delta_{(x_{ij},y_{ij})}$. The
function $\psi$ is a binary classifier that takes a fixed-length mean embedding
as input \citep{Lopez-Paz15b}. 

In kernel methods, $\phi$ is fixed a priori and defined with respect to a
nonlinear kernel \citep{smola2007hilbert}. In contrast, our feature map $\phi :
\mathbb{R}^2 \to \mathbb{R}^h$ and our classifier $\psi : \mathbb{R}^h \to
\{0,1\}$ are both multilayer perceptrons, which are learned jointly from data.
Figure~\ref{fig:ncc} illustrates the proposed architecture, which we term the
Neural Causation Coefficient (NCC). In short, to classify a sample $S_i$ as
causal or anticausal, NCC maps each point $(x_{ij}, y_{ij})$ in the sample
$S_i$ to the representation $\phi(x_{ij},y_{ij}) \in \mathbb{R}^h$, computes
the embedding vector $\phi_{S_i} := \frac{1}{m_i} \sum_{j=1}^{m_i}
\phi(x_{ij},y_{ij})$ across all points $(x_{ij}, y_{ij})\in S_i$, and
classifies the embedding vector $\phi_{S_i} \in \mathbb{R}^h$ as causal or
anticausal using the neural network classifier $\psi$. Importantly, the
proposed neural architecture is not restricted to cause-effect inference, and
can be used to represent and learn from general distributions.

NCC has some attractive properties. First, predicting the cause-effect relation
for a new set of samples at test time can be done efficiently with a single
forward pass through the aggregate network.  The complexity of this operation
is linear in the number of samples.  In contrast, the computational complexity
of the state-of-the-art (kernel-based additive noise models) is cubic in the
number of samples.  Second, NCC can be trained using mixtures of different
causal and anticausal generative models, such as linear, non-linear, noisy, and
deterministic mechanisms linking causes to their effects.  This rich training
allows NCC to learn a diversity of causal footprints simultaneously.  Third,
for differentiable activation functions, NCC is a differentiable function. This
allows us to embed NCC into larger neural architectures or to use it as a
regularization term to encourage the learning of causal or anticausal patterns.

The flexibility of NCC comes at a cost. In practice, labeled cause-effect data
as in Equation~\eqref{eq:causal-data} is scarce and laborious to collect.
Because of this, we follow \cite{Lopez-Paz15b} and train NCC on artificially
generated data.
This turns out to be advantageous as it gives us easy access to unlimited data.
In the following, we describe the process to generate synthetic cause-effect
data along with the training procedure for NCC, and demonstrate the performance
of NCC on real-world cause-effect data.

\subsection{Synthesis of training data}
\label{sec:syntheticdata}

Causal signals differ significantly from the correlation structures exploited
by modern computer vision algorithms. In particular, since the first and second
moments are always symmetrical, causal signals can only be found in high-order
moments.

More specifically, we will construct $n$ synthetic observational samples $S_i$
(see Figure~\ref{fig:models}), where the $i$th observational sample contains
$m_i$ points. The points comprising the observational sample $S_i = \{(x_{ij},
y_{ij})\}_{j=1}^{m_i}$ are drawn from an heteroscedastic additive noise model
$y_{ij} \leftarrow f_i(x_{ij}) + v_{ij}e_{ij}$, for all $j=1, \ldots, m_i$.  In
this manner, we generalize the homoscedastic noise assumption ubiquitous in
previous literature \cite{Mooij14}.

The \emph{cause terms} $x_{ij}$ are drawn from a mixture of $k_i$ Gaussians
distributions. We construct each Gaussian by sampling its mean from
$\text{Gaussian}(0,r_i)$, its standard deviation from $\text{Gaussian}(0,s_i)$
followed by an absolute value, and its unnormalized mixture weight from
$\text{Gaussian}(0,1)$ followed by an absolute value. We sample
$k_i\nsim\text{RandomInteger}[1,5]$ and $r_i, s_i\nsim\text{Uniform}[0,5]$. We
normalize the mixture weights to sum to one. We normalize
$\{x_{ij}\}_{j=1}^{m_i}$ to zero mean and unit variance.

The \emph{mechanism $f_i$} is a cubic Hermite spline with support
\begin{equation}\label{eq:support}
  \begin{aligned}
    \big[~\min(\{x_{ij}\}_{j=1}^{m_i})&-\text{std}(\{x_{ij}\}_{j=1}^{m_i})~ , \\
          \max(\{x_{ij}\}_{j=1}^{m_i})&+\text{std}(\{x_{ij}\}_{j=1}^{m_i}) ~\big]
  \end{aligned}
\end{equation}
and $d_i$ knots drawn from $\text{Gaussian}(0,1)$, 
where $d_i\nsim\text{RandomInteger}(4,5)$. The noiseless effect terms
$\{f(x_{ij})\}_{j=1}^{m_i}$ are normalized to have zero mean and unit variance.

The \emph{noise terms $e_{ij}$} are sampled from
$\text{Gaussian}(0,v_i)$, where $v_i\nsim\text{Uniform}[0,5]$.  To
generalize the ICM, we allow for heteroscedastic noise: we multiply
each $e_{ij}$ by $v_{ij}$, where $v_{ij}$ is the value of a smoothing
spline with support defined as in Equation~\eqref{eq:support} and $d_i$
random knots drawn from $\text{Uniform}[0,5]$.  The noisy effect terms
$\{y_{ij}\}_{j=1}^{m_i}$ are normalized to have zero mean and unit
variance.

This sampling process produces a training set of $2n$ labeled observational samples 
\begin{equation}\label{eq:minibatch}
  \begin{aligned}
    D ~ = ~ & \big\{~(\{(x_{ij},y_{ij})\}_{j=1}^{m_i}, 0)~\big\}_{i=1}^n \\
            & \cup \big\{~(\{(y_{ij},x_{ij})\}_{j=1}^{m_i}, 1)~\big\}_{i=1}^n.
  \end{aligned}
\end{equation}

\subsection{Training NCC}

We train NCC with two embedding layers and two classification layers followed
by a softmax output layer.  Each hidden layer is a composition of batch
normalization \citep{ioffe2015batch}, $100$ hidden neurons, a rectified linear
unit, and $25\%$ dropout \citep{srivastava2014dropout}. We train for $10000$
iterations using RMSProp \citep{rmsprop} with the default parameters, where
each minibatch is of the form given in Equation~\eqref{eq:minibatch} and has
size $2n = 32$. Lastly, we further enforce the symmetry $\mathbb{P}(X \to Y) =
1 - \mathbb{P}(Y \to X)$, by training the composite classifier 
\begin{equation}\label{eq:composite}
  \begin{aligned}
    \tfrac12\big(  1 & - \text{NCC}(\{(x_{ij},y_{ij})\}_{j=1}^{m_i}) \\
                   & + \text{NCC}(\{(y_{ij},x_{ij})\}_{j=1}^{m_{i}}) \big)\,,
  \end{aligned}
\end{equation}
where $\text{NCC}(\{(x_{ij},y_{ij})\}_{j=1}^{m_i})$ tends to zero if the
classifier believes in $X_i \to Y_i$, and tends to one if the classifier
believes in $X_i \leftarrow Y_i$.  We chose our parameters by monitoring the
validation error of NCC on a held-out set of $10000$ synthetic observational
samples. Using this held-out set, we cross-validated the 
dropout rate over $\{0.1,0.25,0.3\}$, the number of hidden layers over $\{2,3\}$,
and the number of hidden units in each of the layers over $\{50,100,500\}$.

\subsection{Testing NCC}

We test the performance of NCC on the T\"ubingen datasets, version 1.0
\citep{Mooij14}.  This is a collection of one hundred heterogeneous,
hand-collected, real-world cause-effect observational samples that are widely
used as a benchmark in the causal inference literature \citep{Lopez-Paz15b}.
The NCC model with the highest synthetic held-out validation accuracy correctly
classifies the cause-effect direction of $79\%$ of the T\"ubingen datasets
observational samples. This result outperforms the previous state-of-the-art on
observational cause-effect discovery, which achieves $75\%$ accuracy on this
dataset \citep{Lopez-Paz15b}.\footnote{The accuracies reported in
\citet{Lopez-Paz15b} are for version 0.8 of the dataset, so we reran the
algorithm from \citet{Lopez-Paz15b} on version 1.0 of the dataset.}

This validation highlights a crucial fact: even when trained on abstract
data, NCC discovers the correct cause-effect relationship in a wide variety of
real-world datasets. But: Do these abstract, domain-independent, causal
footprints hide in complex image data?

\section{Causal signals in sets of static images} \label{sec:visual-causation}

We now have at our disposal all the necessary tools to verify our hypotheses.
In the following, we chose to work with the twenty
object categories of the Pascal VOC 2012 dataset
\cite{pascal-voc-2012}. We first explain how we use NCC to select the
most plausible causal or anticausal features for each object category.
We then we show that the selected anticausal features are more likely
to be object features, that is, located within the object bounding
box, than the selected causal features.  This establishes that
Hypothesis~\ref{hyp:our-hypothesis} is true, and, as a consequence, also
establish that Hypothesis~\ref{hyp:thereissignal} is true.

\subsection{Datasets}

Our experiments use a feature extraction network trained on the
ImageNet \cite{ImageNET} dataset and a classifier network trained on
the Pascal VOC 2012 dataset \cite{pascal-voc-2012}. We then use these
networks to identify causal relationships on the subset of the 99,309
MSCOCO images \cite{lin2014microsoft} representing objects belonging
to the twenty Pascal categories: aeroplane, bicycle, bird, boat,
bottle, bus, car, cat, chair, cow, dining table, dog, horse,
motorbike, person, potted plant, sheep, sofa, train, and
television. These datasets feature heterogeneous images that possibly
contain multiple objects from different categories. The objects may
appear at different scales and angles, and be partially visible or
occluded.  In addition to these challenges, we have no control about
the confounding and selection bias effects polluting these datasets of
images.  All images are rescaled to ensure that their shorter side is
$224$ pixels long, then cropped to the central $224{\times}224$
square.

\subsection{Selecting causal and anticausal features}

Our first task is to determine which of the features scores computed
by the feature extraction neural network represent real world
entities that cause the presence of the object of interest (causal features),
or are caused by the presence of the object of interest (anticausal
features).

To that effect, we consider the feature scores computed by a 18-layer
ResNet \cite{KaimingHeZRS16} trained on the ImageNet dataset
using a proven implementation \cite{resnet}.  Building on top of these
features, we use the Pascal VOC2012 dataset to train
an independent network with two 512-unit hidden layers
to recognize the $20$ Pascal VOC2012 categories, 

For each of the MSCOCO images containing at least one instance of the
twenty Pascal VOC 2012 object categories,
$x_j\in\mathbb{R}^{3{\times}224{\times}224}$, let
$f_j=f(x_j)\in\mathbb{R}^{512}$ denote the vector of feature scores
(before the ReLU nonlinearity) obtained using the feature extraction
network and let $c_j=c(x_j)\in\mathbb{R}^{20}$ denote the vector of
log-odds (that is, the output unit activations before the sigmoid
nonlinearity) obtained using the classifier network. We use features
before their nonlinearity and log odds instead of the class
probabilities because NCC is trained on continuous data with full
support on $\mathbb{R}$. 

As depicted in Figure~\ref{fig:aliasing}, for each category $k\in\{1\dots20\}$
and each feature $l\in\{1\dots512\}$, we apply NCC to the scatterplot
$\{(f_{jl},c_{jk})\}_{j=1}^m$ representing the joint distribution of
the scores of feature $j$ and the score of category $k$.
Since these scores are computed by running our neural networks
on the image pixels, they are not related by a direct causal relationship.
However we know that these scores are highly correlated with the presence
of objects and features in the real scene. Therefore, the appearance
of a causal relationship between these scores suggests that there
is a causal relationship between the real world entities they represent.

Because we analyze one feature at a time, the values taken by all
other features appear as an additional source of noise, and the
observed statistical dependencies are then be much weaker than in the
synthetic NCC training data. To avoid detecting causation between
independent random variables, we use a variant of NCC trained with an
augmented training set: in addition to presenting each scatterplot in
both causal directions as in \eqref{eq:minibatch}, we pick a random
permutation $\sigma$ to generate an additional uncorrelated example
$\{x_{i,\sigma(j)},y_{ij}\}_{j=1}^{m_i}$ with label $\tfrac{1}{2}$. We
use our best model of this kind which, for validation purposes,
achieves $79\%$ accuracy in the T\"ubingen pair benchmark.

For each category $k\in{1\dots20}$, we then record the indices of
the top 1\% causal and the top 1\% anticausal features.

\subsection{Verifying Hypothesis~\ref{hyp:our-hypothesis}}

\begin{figure}[t]
  \begin{center}
  \begin{subfigure}{.32\linewidth}
    \centering
    \includegraphics[width=\textwidth]{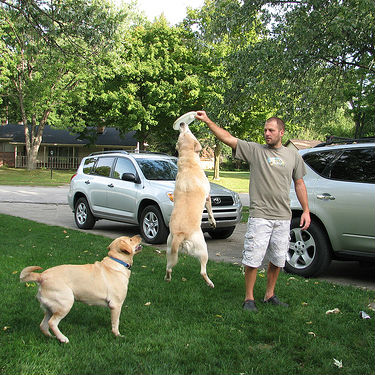}
    \caption{$x_j$}
    \label{fig:images:original}
  \end{subfigure}
  \hfill
  \begin{subfigure}{.32\linewidth}
    \centering
    \includegraphics[width=\textwidth]{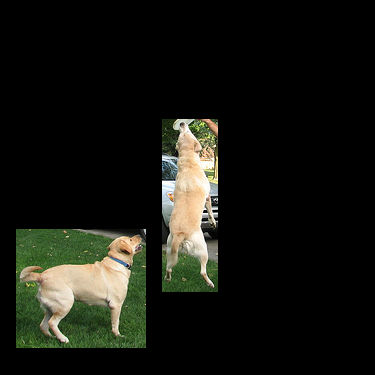}
    \caption{$x^o_j$}
    \label{fig:images:object}
  \end{subfigure}
  \hfill
  \begin{subfigure}{.32\linewidth}
    \centering
    \includegraphics[width=\textwidth]{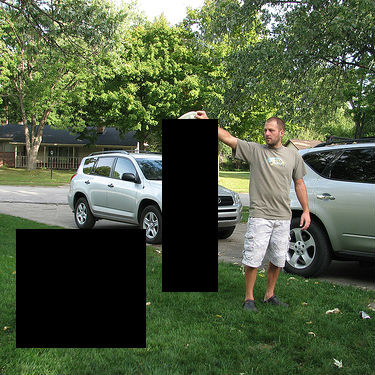}
    \caption{$x^c_j$}
    \label{fig:images:context}
  \end{subfigure}
  \end{center}
  \vspace*{-2ex}
  \caption{Blacking out image pixels to distinguish object-features
    from context-features. We show the original image~$x_j$, and the
    corresponding object-only image $x^o_j$ and context-only image
    $x^c_j$ for the category ``dog''. The pixels are blacked out after
    normalizing the image in order to obtain true zero pixels.}
  \label{fig:images}
\end{figure}

\begin{figure*}[t]
  \begin{center}
  \includegraphics[width=\linewidth]{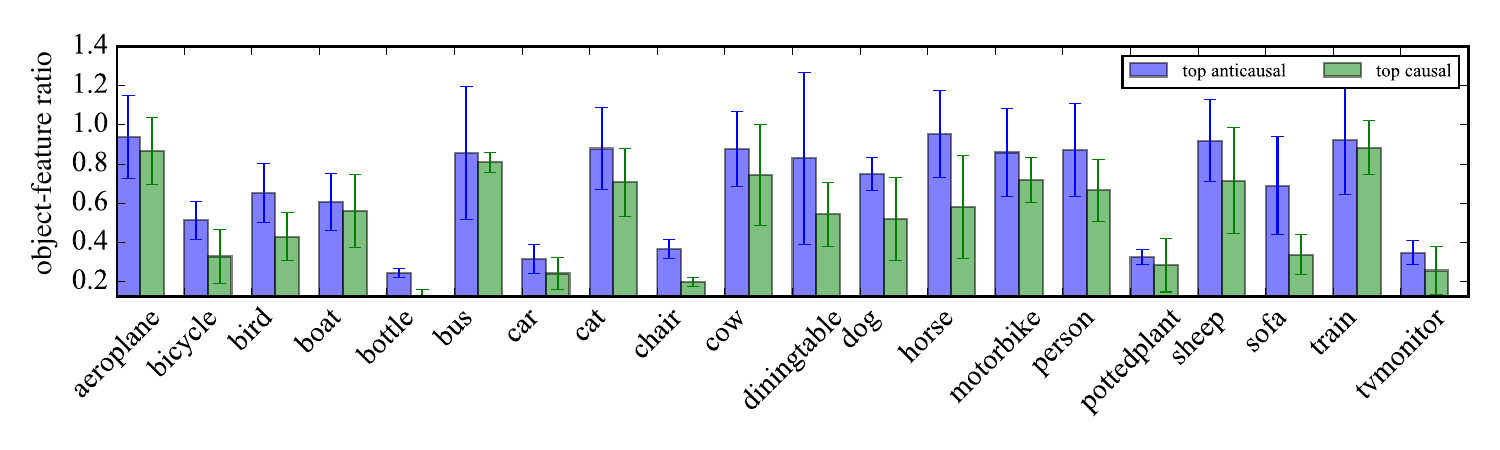}
  \includegraphics[width=\textwidth]{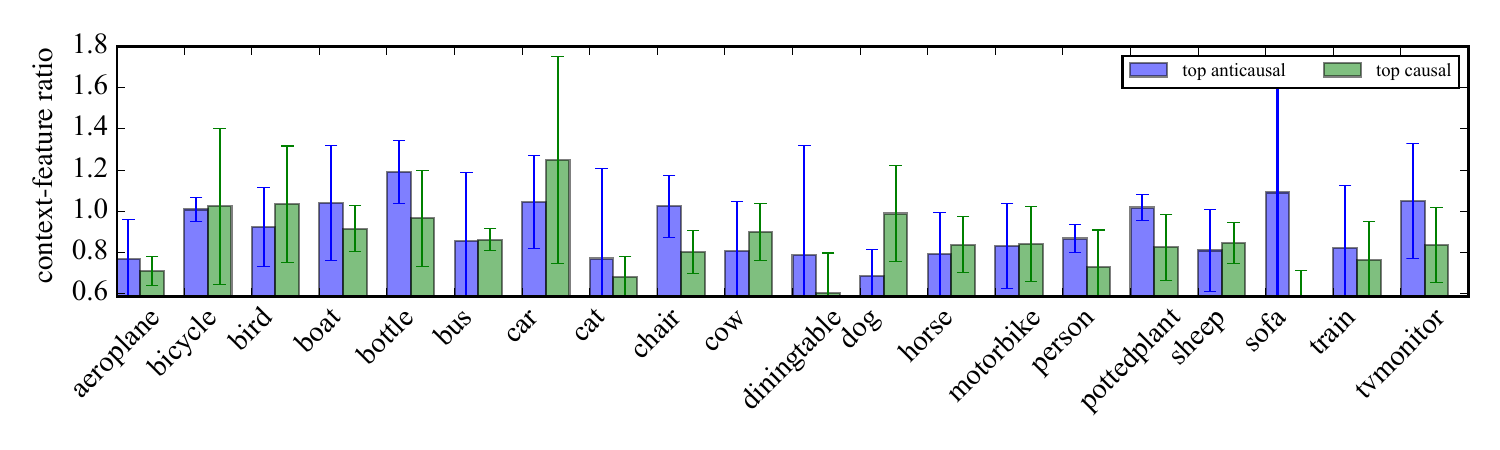}
  \end{center}
  \vspace{-0.5cm}
  \caption{Average and standard deviation of the object/context feature scores
  associated to the top $1\%$ causal/anticausal feature scores, for all the
  twenty studied categories. The average object feature score associated to the
  top $1\%$ anticausal feature scores is always higher than the average object
  feature score associated to the top $1\%$ causal features. Such separation
  does not occur for context feature scores. These results are strong empirical
  evidence in favour of Hypoteheses~\ref{hyp:thereissignal} and
  \ref{hyp:our-hypothesis}: the probability of obtaining these results by
  chance is $2^{-20}\approx10^{-6}$.}
  \label{fig:results}
\end{figure*}

In order to verify Hypothesis~\ref{hyp:our-hypothesis}, it is
sufficient to show that the top anticausal features are more likely to
be object features than the top causal features. For each category $k$
and each feature $j$, we must therefore determine whether feature $j$
is likely to be an object feature or a context feature.  This is
relatively easy because we have access to the object bounding boxes
and we simply need to determine how much of each feature score $j$ is
imputable to the bounding boxes of the objects of category $k$.

To that effect, we prepare two alternate versions of each MSCOCO image
$x_j$ by blacking out (with zeroes) the pixels located outside the
bounding boxes of the category $k$ objects, yielding the object-only
image $x^o_j$, or by blacking out the pixels located inside the
bounding boxes of the category $k$ objects, yielding the context-only
image $x^c_j$.  This process is illustrated in
Figure~\ref{fig:images:context}.  We then compute the corresponding
vectors of feature scores $f^o_j=f(x^o_j)$ and $f^c_j=f(x^c_j)$.

For each category $k$ and each feature $f$ we heuristically define the
\emph{object-feature ratio} $s^o_l$ and the \emph{context-feature ratio}
$s^c_l$ as follows:
\[
s^o_l = \frac{\sum_{j=1}^m \left| f^c_{jl} - f_{jl} \right|}
             {\sum_{j=1}^m \left| f_{jl} \right|}~,
\quad
s^c_l = \frac{\sum_{j=1}^m \left| f^o_{jl} - f_{jl} \right|}
              {\sum_{j=1}^m \left| f_{jl} \right|}~.
\]
Intuitively, features with high object-feature ratio (resp. high
context-feature ratio) are those features that react violently when
the object (resp. the context) is erased.

Note that blacking out pixels does \emph{not} constitute an
intervention on the scene represented by the image. This is merely a
procedure to impute the contribution of the object bounding boxes to
each feature score.

\subsection{Results}

Figure~\ref{fig:results} shows the means and the standard deviations of the
object-context ratios (top plot) and the context-feature ratios (bottom plot)
estimated on the top 1\% anticausal features (blue bars) and the top 1\% causal
features (green bars) for each of the twenty object categories.

As predicted by Hypothesis~\ref{hyp:our-hypothesis}, object features are
related to anticausal features: the top 1\% anticausal features exhibit a
higher object-feature ratio than the top 1\% causal features. Since this effect
can be observed on all $20$ classes of interest, the probability of obtaining
such a result by chance would be $2^{-20}{\approx}10^{-6}$. When we select the
top 20\% causal and anticausal features, this effect remains consistent across
$16$ out of $20$ classes of interest.

This result indicates that anticausal features may be useful for detecting
objects locations in a robust manner, regardless of their context. As stated in
Hypothesis~\ref{hyp:our-hypothesis}, we could not find a consistent
relationship between context features and causal features. Remarkably, we
remind the reader that the NCC classifier \emph{does not depend on the object
categories} and was \emph{trained using synthetic data unrelated to images}.
As a sanity check, we did not obtain any such results when replacing the NCC
scores with the correlation coefficient or the absolute value of the
correlation coefficient.\footnote{\relax We also ran preliminary experiments to
find causal relationships between objects of interest, by computing the NCC
scores between the log odds of different objects of interest.  The strongest
causal relationships that we found were ``bus causes car,'' ``chair causes
plant,'' ``chair causes sofa,'' ``dining table causes bottle,'' ``dining table
causes chair,'' ``dining table causes plant,'' ``television causes chair,'' and
``television causes sofa.''}

Therefore we believe that this result establishes that
Hypothesis~\ref{hyp:our-hypothesis} is true with high certainty. As explained
in Section~\ref{sec:our-contribution}, verifying
Hypothesis~\ref{hyp:our-hypothesis} in this manner also implies confirms
Hypothesis~\ref{hyp:thereissignal}.

\section{Conclusion}
\label{sec:conclusion}

Using a carefully designed experiment, we have established that the
high order statistical properties of image datasets contain
information about the causal dispositions of objects and, more
generally, about causal structure of the real world.

Our experiment relies on three main components. First, we use synthetic
scatterplots to train a binary classifier that identifies plausible
causal ($X{\rightarrow}Y$) and anticausal ($X{\leftarrow}Y$)
relations. Second we hypothesise that the distinction between object
features and context features in natural scenes is related to the
distinction between features that cause the presence of the object and
features that are caused by the presence of the object.  Finally,
we construct an experiment that leverages static image datasets to 
establish that this latter hypothesis is true. Thus, we conclude
that we must therefore have been able to effectively
distinguish which features were causal or anticausal.

Because we now know that such a signal exist, we can envision in a
reasonable future that computer vision algorithms will be able to
perceive the causal structure of the real world and reason about scenes.
There is no question that significant algorithmic advances will be
necessary to achieve this goal. In particular, we stress the
importance of (1) building large, real-world datasets to aid research
in causal inference, (2) extending data-driven techniques like NCC to
causal inference of more than two variables, and (3) exploring data
with explicit causal signals, such as the arrow of time in videos
(\eg \cite{PickupPWSZZSF14}.)
  
\clearpage
\newpage
{\small
\bibliographystyle{ieee}
\bibliography{viscaus}
}

\end{document}